%% file: main.tex
\documentclass[10pt,twocolumn,letterpaper]{article}

\usepackage{cvpr}              
\usepackage{amsmath}
\usepackage{multirow}
\usepackage{pifont}
\usepackage{soul}
\input{preamble}

\definecolor{cvprblue}{rgb}{0.21,0.49,0.74}
\usepackage[pagebackref,breaklinks,colorlinks,allcolors=cvprblue]{hyperref}

\def\paperID{****} 
\def\confName{CVPR}
\def\confYear{2025}

\title{Reconstructing Close Human Interaction with Appearance and Proxemics Reasoning}
\author{Buzhen Huang$^{1,2}$\footnotemark[1]\hspace{5mm} Chen Li$^{4,5}$\hspace{5mm} Chongyang Xu$^{3}$\hspace{5mm} Dongyue Lu$^{2}$\hspace{5mm} Jinnan Chen$^{2}$\hspace{5mm} \\
Yangang Wang$^{1}$\hspace{5mm} Gim Hee Lee$^{2}$\\%
\\
$^1$Southeast University \hspace{1mm}
$^2$National University of Singapore \hspace{1mm} 
$^3$Sichuan University \hspace{1mm}\\
$^4$IHPC, Agency for Science, Technology and Research, Singapore\\
$^5$CFAR, Agency for Science, Technology and Research, Singapore\\
\\
\vspace{-13mm}
}

\begin{document}

\twocolumn[{%
\renewcommand\twocolumn[2][]{#1}%
\maketitle
\thispagestyle{empty}
\vspace{-7mm}
\begin{center}
   \centering
   \includegraphics[width=0.95\textwidth]{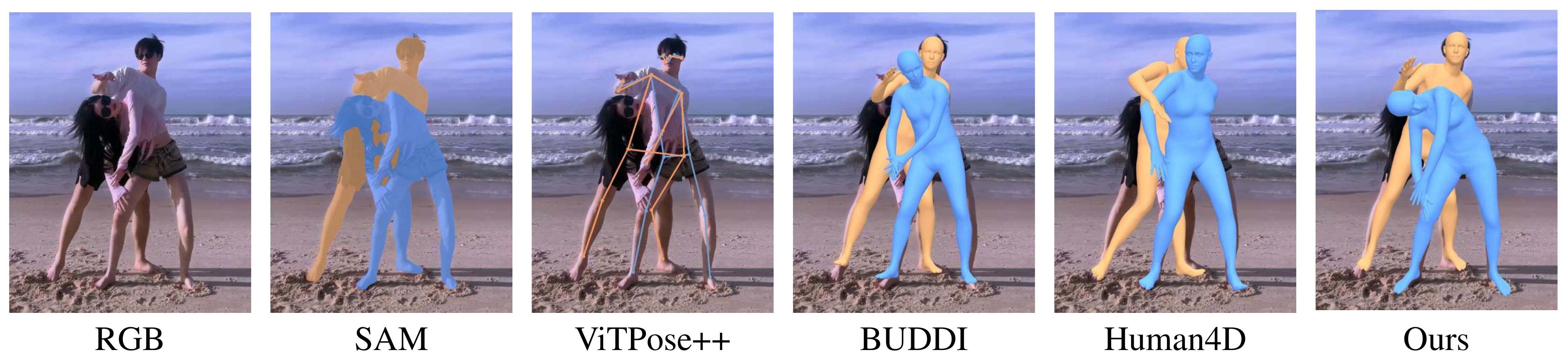}
   \vspace{-3mm}
   \captionof{figure}{Due to the visual ambiguity, even state-of-the-art vision foundation models~(\eg, ViTPose++~\cite{xu2023vitpose} and temporal SAM~\cite{kirillov2023segment,AutoTrackAnything}) cannot clearly distinguish human semantics in close interactive cases. Consequently, human pose estimation methods based these basic human semantics tend to fail. In comparison, our dual-branch optimization framework that leverages human appearance, proxemics, and physics is capable of alleviating visual ambiguity to give better results.}
   \label{fig:teaser}
\end{center}
}]

\footnotetext[1]{The work was done while Buzhen Huang is a visiting student at National University of Singapore.}

\input{sec/0_abstract}    
\input{sec/1_intro}

\input{sec/2_related_work}

\input{sec/3_method}
\input{sec/4_dataset}

\input{sec/5_experiments}
\input{sec/6_conclusion}

{
    \small
    \bibliographystyle{ieeenat_fullname}
    \bibliography{main}
}


\end{document}

%% file: preamble.tex
\DeclareMathOperator*{\argmin}{arg\,min}


%% file: sec/0_abstract.tex
\begin{abstract}
    Due to visual ambiguities and inter-person occlusions, existing human pose estimation methods cannot recover plausible close interactions from in-the-wild videos. Even state-of-the-art large foundation models~(\eg, SAM) cannot accurately distinguish human semantics in such challenging scenarios. In this work, we find that human appearance can provide a straightforward cue to address these obstacles. Based on this observation, we propose a dual-branch optimization framework to reconstruct accurate interactive motions with plausible body contacts constrained by human appearances, social proxemics, and physical laws. Specifically, we first train a diffusion model to learn the human proxemic behavior and pose prior knowledge. The trained network and two optimizable tensors are then incorporated into a dual-branch optimization framework to reconstruct human motions and appearances. Several constraints based on 3D Gaussians, 2D keypoints, and mesh penetrations are also designed to assist the optimization. With the proxemics prior and diverse constraints, our method is capable of estimating accurate interactions from in-the-wild videos captured in complex environments. We further build a dataset with pseudo ground-truth interaction annotations, which may promote future research on pose estimation and human behavior understanding. Experimental results on several benchmarks demonstrate that our method outperforms existing approaches. The code and data are available at \url{https://www.buzhenhuang.com/works/CloseApp.html}. 
\end{abstract}

%% file: sec/1_intro.tex
\section{Introduction}\label{sec:Introduction}
Human interaction is an essential part of life and has many physical, mental and emotional benefits. Enabling machines to understand human interaction may promote a lot of downstream applications such as robotics, virtual reality, smart security, \textit{etc}. As an effective tool for understanding human behaviour, 3D human pose and shape estimation has achieved profound progress in recent years. However, the existing methods are still not deployable for analysing close human interactions due to severe visual ambiguities and inter-person occlusions.

\begin{figure*}
    \begin{center}
    \includegraphics[width=0.8\linewidth]{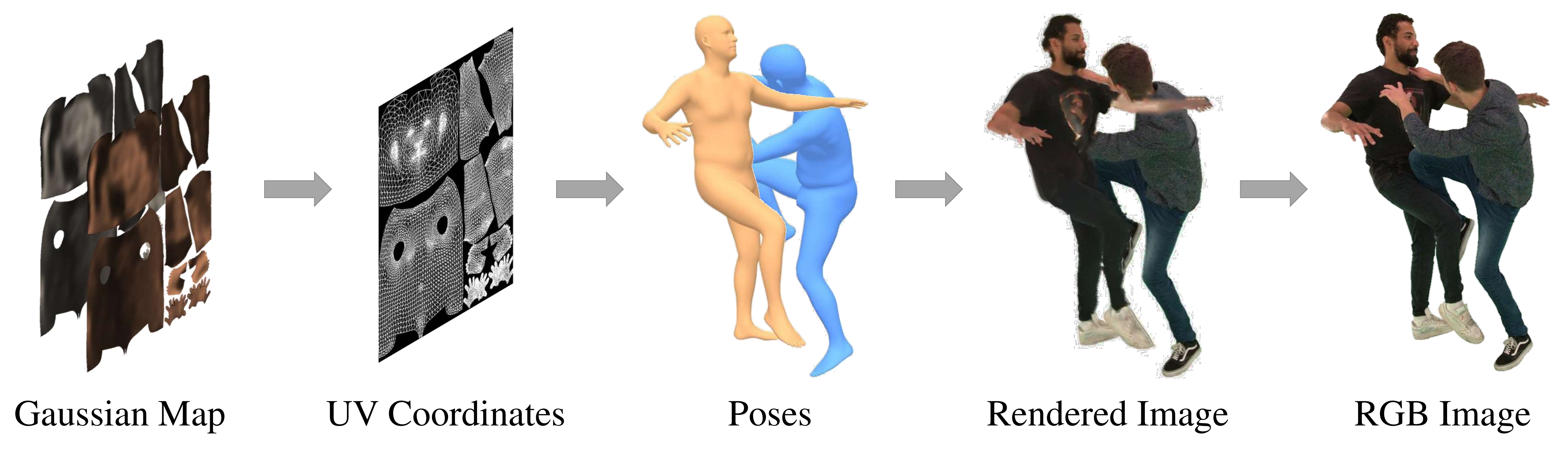}
    \end{center}
    \vspace{-7mm}
    \caption{With predicted UV Gaussian maps, we can map the Gaussians to 3D space with a UV coordinate map and splat them to the image plane. We can then reason the depth ordinal relationship and image-model alignment with the rendered and original images. Since the Gaussians should also be consistent across non-occluded frames, the optimization adjusts poses to find an optimal solution in interactive frames, thereby producing accurate depth ordering and poses.}
\label{fig:appearance}
\vspace{-5mm}
\end{figure*}

Specifically, single-person methods~\cite{stathopoulos2024score,goel2023humans,li2022cliff} only focus on pose accuracy and image-model alignment, while multi-person approaches~\cite{jiang2020coherent,zanfir2018monocular,sun2022putting,huang2023reconstructing} tend to address penetration~\cite{jiang2020coherent,zanfir2018monocular} and spatial distribution reasonableness~\cite{sun2022putting,huang2023reconstructing}. They all ignore the important body contacts and proxemics in close interactions. Only a few recent works are designed for close human-human interactions~\cite{huang2024closely,muller2023generative,ugrinovic2024multiphys,fang2024dualhuman}, which regularize the interactions with learned reaction priors~\cite{huang2024closely,fang2024dualhuman} or physical simulation~\cite{ugrinovic2024multiphys}. However, the regression-based methods~\cite{huang2024closely,ugrinovic2024multiphys,fang2024dualhuman} rely on high-quality interaction data captured in indoor scenes, and thus show poor generalization ability on in-the-wild images. In contrast, BUDDI~\cite{muller2023generative} fits human models to 2D keypoints via an optimization framework and can work in diverse environments. 
Nonetheless, even state-of-the-art large foundation models~(\eg, SAM~\cite{kirillov2023segment} and ViTPose++~\cite{xu2023vitpose}) cannot clearly identify human semantics in complex interaction cases due to the visual ambiguity. Consequently, BUDDI still tend to fail.

In this work, we find that human appearances can provide straightforward cues to alleviate visual ambiguities and inter-person occlusions. As shown in \cref{fig:appearance}, with the modeled human appearances and rendering techniques~\cite{kerbl20233d}, we can directly leverage the original RGB image to infer the depth ordinal relationships and image-model alignment for occluded cases. Based on this observation, we design a novel dual-branch optimization framework constrained by appearance, proxemics and physics to reconstruct close human interactions. 
However, this straightforward idea requires simultaneously reconstructing human motions and appearances, which is a highly non-convex optimization aggravated by depth ambiguity.

To this end, we first propose a diffusion model to learn proxemic behaviors. In contrast to existing interaction priors~\cite{muller2023generative,huang2024closely,fang2024dualhuman}, our model receives 2D observations and infers 3D interactions from both temporal and reactive information. After the training on interactive data with a mask strategy, the model can regularize the interaction from a noisy and partially observed input. Subsequently, the diffusion model with trained parameters is used as the motion branch in the optimization framework. During the optimization, the prior can produce a desired interaction by fine-tuning the network parameters, and this strategy is more robust to depth ambiguity and local minima. We then design an appearance branch with two optimizable tensors to constrain the reconstructed motions. Specifically, the tensors are decoded to Gaussian UV maps with a U-Net~\cite{ronneberger2015u} backbone. The Gaussians are then mapped to 3D body surface with a UV coordinate map. By splatting the Gaussians, we can simultaneously optimize the motions and appearances. In addition, we also penalize mesh penetration and keypoints re-projection error to improve physical plausibility and pose accuracy. 

Since our method can work well on different environments, we further collect 100 human-human interaction videos from Internet and build pseudo ground-truth interaction annotations. Experimental results show that the proposed dataset can improve the current regression-based method, and may promote future research on human interaction understanding. To summarize, the main contributions of this paper are as follows:

\begin{itemize}
  \item We propose a dual-branch optimization framework constrained by appearance, proxemics and physics to reconstruct close human interactions, which can work well on in-the-wild videos.

  \item We demonstrate that human appearance can be an effective cue to alleviate visual ambiguities and inter-person occlusions in closely interactive scenarios.

  \item We build an in-the-wild dataset for close human interactions, which may promote future research on human interaction understanding.
  
\end{itemize}

%% file: sec/2_related_work.tex
\section{Related Work}\label{sec:Related}
\noindent\textbf{Human interaction reconstruction.}
Human pose and shape estimation has made tremendous progress in the past several years. However, most of works~\cite{goel2023humans,stathopoulos2024score,kanazawa2018end,li2022cliff} in this field focus on pose accuracy and image-model alignment for a single person, and ignore the important interactions between humans. Some works~\cite{qiu2023psvt,li2023coordinate,sun2021monocular,choi2022learning} consider multiple humans in the same scene, but they just address the inter-person occlusions and do not reconstruct absolute positions for human interactions. Although some multi-person methods can regress an approximate absolute translation for each human with projection geometry~\cite{cha2022multi,zanfir2018deep,huang2022pose2uv,zanfir2018monocular}, novel position representations~\cite{sun2022putting,zhang2021body}, or ordering-aware loss~\cite{jiang2020coherent,khirodkar2022occluded,huang2023reconstructing,wen2023crowd3d}, the coarse estimation is inadequate for delicate close human interactions. Reconstructing close human interaction is an open issue for decades due to the depth ambiguity, mesh penetration, and inter-person occlusion. Only a few recent works explicitly consider this problem by incorporating collision avoidance~\cite{fieraru2020three,ugrinovic2024multiphys}, contact constraints~\cite{fieraru2023reconstructing,subramanian2024pose}, or proxemics priors~\cite{muller2023generative,huang2024closely,fang2024dualhuman}. However, they all rely on detected 2D human semantics and are still confronted with visual ambiguities. 

\vspace{1mm}
\noindent\textbf{Human Gaussian splatting.}
3D Gaussian Splatting~\cite{kerbl20233d} uses a set of 3D Gaussians to represent a scene and renders it by splatting and rasterizing the Gaussians, which has shown high efficiency and impressive performance on static objects. Recent works have introduced this technique to model dynamic 3D humans~\cite{moreau2023human} and articulated objects~\cite{lei2023gart}. Typically, a predicted human motion with SMPL representation~\cite{loper2015smpl} is used to initialize the Gaussians, and then the properties of each Gaussian is optimized by a rendering-and-compare strategy~\cite{moreau2023human,lei2023gart,qian20233dgs}. To model diverse cloth topologies, some methods adopt a more complex mesh template for the initialization~\cite{jiang2023hifi4g,pang2023ash,li2023animatable} or directly use depth information as input~\cite{zheng2023gps}. Since these methods all rely on multi-view inputs, a few works~\cite{kocabas2023hugs,hu2023gauhuman,hu2023gaussianavatar,wen2024gomavatar} further simplify the settings to use a monocular video to learn the human Gaussians. However, they still require the video to capture the complete observations of a human body. To address the occluded and invisible parts, OccGaussian~\cite{ye2024occgaussian} designs an occlusion feature query to reconstruct humans from partial observations. Lee~\etal~\cite{lee2024guess} also considered the similar obstacle in multi-person scenarios and used a 2D diffusion model to provide additional information. Nonetheless, they iteratively processed each human and still need accurate foreground masks. In contrast to these existing works, we simultaneously predict Gaussians for two characters, and use the reconstructed appearances to constrain the human motions.

\vspace{1mm}
\noindent\textbf{Pseudo ground-truth generation.}
Historically, building 3D human annotations always relies on expensive marker-based~\cite{h36m_pami,xu2023inter,fieraru2020three} or multi-view~\cite{hanbyuljoo2019panoptic,liang2023intergen} systems in controlled environments. Although the captured poses are accurate, the model trained with this data shows poor generalization ability in in-the-wild scenarios due to its simple background and appearance. To close the domain gap, a few methods~\cite{kanazawa2018end,Bogo:ECCV:2016,SMPL-X:2019} have leveraged weak supervision for human reconstruction, but the results are still unsatisfactory due to the sparse constraint. Recently, pseudo ground-truth annotations~\cite{joo2021exemplar,lassner2017unite,pavlakos2022human} have enabled human pose and shape estimation to show impressive performance~\cite{goel2023humans,stathopoulos2024score}. With learned pose prior knowledge, some annotators~\cite{joo2021exemplar,moon2022neuralannot,lin2023one} directly optimize the predictions by finetuning the network parameters. Additional constraints from camera perspective models~\cite{li2022cliff}, temporal dependencies~\cite{pavlakos2022human}, and crowd spatial distribution~\cite{huang2023reconstructing} are also incorporated to improve the annotation quality. Compared to common single- or multi-person data, close human-human interactions are more difficult to obtain. A very recent work, BUDDI~\cite{muller2023generative}, has designed a proxemics prior to generate interaction annotations. However, due to the visual ambiguity and the lack of temporal information, it is still struggle to produce high quality data in severely occluded cases.

%% file: sec/3_method.tex
\section{Our Method}\label{sec:Method}
\begin{figure*}
    \begin{center}
    \includegraphics[width=1.0\linewidth]{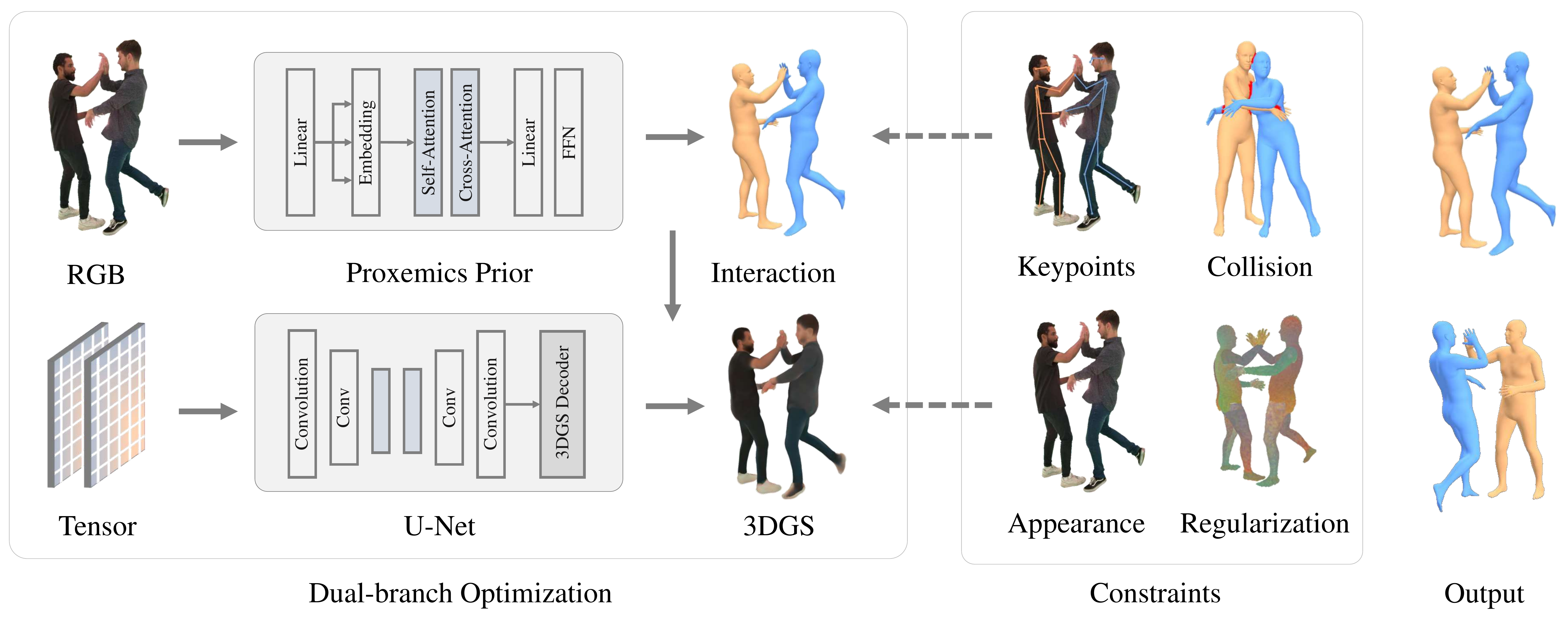}
    \end{center}
    \vspace{-7mm}
    \caption{\textbf{Overview of our framework.} We propose a dual-branch optimization framework to reconstruct close human interactions from a monocular in-the-wild video. By optimizing the proxemics prior, U-Net backbone, and two optimizable tensors, the framework simultaneously predicts interactive motions and coarse appearances. With the constraints from 2D observations, physics, and prior knowledge, the framework can finally output 3D interactions with plausible body poses, natural proxemic relationships and accurate physical contacts.}
\label{fig:framework}
\vspace{-5mm}
\end{figure*}

Fig.~\ref{fig:framework} shows an illustration of our framework. Given a monocular in-the-wild video with close interactions between two people, we propose a dual-branch optimization framework to reconstruct accurate body poses, natural proxemic relationships, and plausible physical contacts. 

\subsection{Human interaction representation}\label{sec:Representation}
\noindent\textbf{Motion representation.} We adopt the SMPL model~\cite{loper2015smpl} with a 6D rotation representation~\cite{zhou2019continuity} to describe the interaction, which consists of pose $\theta \in \mathbb{R}^{144}$, shape $\beta \in \mathbb{R}^{10}$, and translation $\tau \in \mathbb{R}^{3}$. For a video with $N$ frames and 2 individuals, the reconstructed motions can be denoted as $\mathbf{x}^{1:N} = \left\{\mathbf{x}^{a,1:N}, \mathbf{x}^{b,1:N} \right\}$, where $\mathbf{x}^{a,1:N} = \left\{\theta^i, \beta^i, \tau^i\right\}_{i=1}^N$. 

\vspace{1mm}
\noindent\textbf{Appearance representation.} 3D Gaussian Splatting~\cite{kerbl20233d} is used to represent human appearance, which is parameterized by a set of 3D Gaussians. Conventionally, each Gaussian contains an offset relative to SMPL vertex $\mu \in \mathbb{R}^{3}$, color $c \in \mathbb{R}^{3}$, opacity $\sigma \in \mathbb{R}$, rotation $q \in \mathbb{R}^{3}$, and scale $s \in \mathbb{R}^{3}$. We can obtain the rendered appearance by splatting Gaussians to the image plane.

\subsection{Proxemic prior}\label{sec:prior}
Directly optimizing SMPL parameters~\cite{bogo2016keep} for close human reconstruction encounters severe depth ambiguity and is highly sensitive to occlusions and local minima. To address these obstacles, we first train a diffusion model to learn pose and proxemics prior knowledge to assist the optimization.

\vspace{1mm}
\noindent\textbf{Model architecture.}
We adopt a diffusion model to learn the prior, which iteratively predicts clean data from a pure noise conditioned on 2D observations. In addition to image features, we use 2D keypoints as an additional condition for the diffusion since the existing interaction datasets~\cite{liang2023intergen,xu2023inter} may not contain paired RGB images. When the RGB image is unavailable, we can still use these data to train the prior by setting image features to be zero. Specifically, the ground-truth two-person motions $\hat{\mathbf{x}}_0^{1:N}$ are first diffused towards a standard Gaussian distribution:
\begin{equation}\label{equation:forward_diffusion_origin}
    q(\mathbf{x}_t \mid \hat{\mathbf{x}}_0) = \sqrt{\hat{\alpha}_t}\hat{\mathbf{x}}_0 + \sqrt{1 - \hat{\alpha}_t} \epsilon, \epsilon \sim \mathcal{N} \left(0, \rm{I} \right),
\end{equation}
where $\alpha_t$ and $\hat{\alpha}_t$ are constant hyper-parameters~\cite{nichol2021improved}. The noisy motions $\mathbf{x}_t^{1:N}$ are projected to high-dimensional vectors and then concatenated with image and keypoints features. We omit illustration of the full diffusion process in Fig.~\ref{fig:framework} for brevity. $S$ transformer blocks are used to process the concatenated features. 
As shown in the top left of Fig.~\ref{fig:framework}, the features of two individuals can share information with a cross-attention module~\cite{liang2023intergen} in each transformer block.
Finally, the denoised motions are regressed with a feed-forward layer from the processed features. In each timestep, the diffusion model predicts the clean motions and then diffuses them to $\mathbf{x}_{t-1}^{1:N}$, which is defined as:
\begin{equation}\label{equation:reverse_diffusion}
    q(\mathbf{x}_{t-1} \mid \mathbf{x}_{t},c) = \mathcal{N} (\mathbf{x}_{t-1}; \mu_{\alpha}(\mathbf{x}_{t},c),\tilde{\gamma_t} \rm{I}),
\end{equation}
where $\mu_{\alpha}(\mathbf{x}_{t},c)$ is the estimated mean by the diffusion model under the condition of $c$. $\tilde{\gamma_t}$ is a 
hyper-parameter.

\vspace{1mm}
\noindent\textbf{Mask strategy.}
To ensure that the prior is robust to occlusions, we adopt two mask strategies to learn temporal dependencies and proxemic behaviors, respectively. 
1) We randomly mask the condition and input poses 
from a subset of frames, and then enforce the diffusion model to inpaint the missing information based on temporal relationships. 
2) We may completely mask the inputs of one individual and compel the model to generate a reaction from the counterpart. With these two strategies, the prior can produce complete motions even when one individual is totally invisible.

\vspace{1mm}
\noindent\textbf{Training loss.}
We use the following loss functions to train the prior: 
\begin{equation}
    \mathcal{L} = \mathcal{L}_{\text{reproj}} + \mathcal{L}_{\text{smpl}} + \mathcal{L}_{\text{joint}} + \mathcal{L}_{\text{vel}} + \mathcal{L}_{\text{int}}.
\end{equation}\label{equ:prior}
The reprojection loss is given by:
\begin{equation}
    \mathcal{L}_{\text{reproj}} = \| \Pi\left(J_{3 D}+ \tau\right) - \hat{J}_{2D} \|_2^2,\label{equ:reproj}
\end{equation}
where $\hat{J}_{2D}$ is ground-truth 2D pose. It should be noted that the reprojection loss is important for regressing the absolute translations. Previous single-person methods~\cite{stathopoulos2024score,goel2023humans} always use a weak-perspective camera and may result in unreasonable spatial distribution in interaction reconstruction. 
We thus follow CLIFF~\cite{li2022cliff} to use a perspective camera $\Pi(\cdot)$ with a common diagonal Field-of-View $55^\circ$ to project the 3D joints to 2D image. The remaining terms include the supervisions from the SMPL parameters: $\mathcal{L}_{\text{smpl}} = \| [\beta, \theta] - [\hat{\beta}, \hat{\theta}] \|_2^2$, 3D joint positions: $\mathcal{L}_{\text{joint}} = \| J_{3D} - \hat{J_{3D}} \|_2^2$, and velocities: $\mathcal{L}_{\text{vel}} = \| \dot{J}_{3D} - \hat{\dot{J}}_{3D} \|_2^2$ on each individual. To enforce more plausible interaction, we also penalize the relative distance between two characters:
\begin{equation}
    \mathcal{L}_{\text{int}} = \| |J^a_{3D} - J^b_{3D}| - |\hat{J}^a_{3D} - \hat{J}^b_{3D}| \|_2^2.
\end{equation}

Once the training is completed, the diffusion model can predict close interactions from RGB images and 2D keypoints. However, due to the visual ambiguity and occlusion, the results may not be consistent with image observations. 
To mitigate this issue, we use the predicted motions and trained network parameters as initial values, and further refine the motions with a dual-branch optimization.

\subsection{Dual-branch optimization}\label{sec:optimization}
Due to the visual ambiguity, the current regression models~\cite{xu2023vitpose,kirillov2023segment} cannot clearly distinguish human semantics in closely interactive cases, and thus feed-forward human reconstruction tends to fail. We therefore design a dual-branch optimization to leverage human appearance, proxemics, and physics to address these problems. 

\vspace{1mm}
\noindent\textbf{Motion branch.} 
We utilize the trained diffusion model from the previous section to construct the motion branch. Initially, motions $\mathbf{x}_0^{1:N}$ are regressed from image and keypoints features. As these motions may exhibit image-model misalignment and lack reliability, we then diffuse them to $\mathbf{x}_1^{1:N}$. Unlike traditional reverse diffusion processes that adjust motions under additional guidance~\cite{stathopoulos2024score,huang2024closely}, we finetune the network parameters $\pi_{m}$ using several loss functions to update $\mathbf{x^\prime}_0^{1:N}$. This approach enhances the controllability of reconstructed motions and ensures their consistency with observations. Moreover, the pretrained network parameters can provide pose and proxemic prior knowledge to alleviate the 
depth ambiguity and occlusion.

\vspace{1mm}
\noindent\textbf{Appearance branch.} 
Previous optimization-based methods iteratively fit the model to 2D measurements like 2D keypoints~\cite{bogo2016keep,SMPL-X:2019}, silhouette~\cite{wang2017outdoor}, or part segmentation~\cite{lassner2017unite}. However, even state-of-the-art large foundation models~\cite{kirillov2023segment,xu2023vitpose} struggle to produce accurate human semantics for close human interactions due to visual ambiguity. We find that RGB images can provide reliable dense correspondences and can serve as a constraint with a reconstructed human appearance. 
Consequently, we design the appearance branch to predict 3D Gaussians for appearance modeling as shown in the bottom left of Fig.~\ref{fig:framework}. 
The Gaussian properties are encoded in a UV map~\cite{jiang2024uv} since it is difficult to directly optimize tens of thousands of independent Gaussians.
We use a U-Net backbone to regress the map from an optimizable tensor 
which works as a latent code for the human appearance. The Gaussian UV map has 14 channels containing offset $\mu$, color $c$, opacity $\sigma$, rotation $q$, scale $s$, and identity $d$. The convolution layers build the dependencies for Gaussians on the UV map, and each Gaussian can be mapped to 3D space with the UV coordinate map. By optimizing the input tensors and U-Net, we can reconstruct the human appearances. With the poses from the motion branch, we can also render the appearances to 2D images via Gaussian splatting. 

\vspace{1mm}
\noindent\textbf{Objective function.}
We formulate several objective functions to constrain the output of the dual-branch framework. We first penalize the appearance loss:
\begin{equation}
    \mathcal{L}_{\text{app}} = \mathcal{L}_{\text{rgb}} + \mathcal{L}_{\text{ssim}} + \mathcal{L}_{\text{lpips}},
\end{equation}
where $\mathcal{L}_{\text{rgb}}$, $\mathcal{L}_{\text{ssim}}$, and $\mathcal{L}_{\text{lpips}}$ are the L1, SSIM~\cite{wang2004image}, and LPIPS~\cite{zhang2018unreasonable} loss between rendered and original images. We also calculate the re-projection loss~\cref{equ:reproj} with 2D keypoints to prevent large pose deviations. Since we can access the region of the Gaussians rendered on the image with identity $d$, we only use the keypoints that fall within the rendered region of each individual, which differs from the common formulation and helps alleviate the impact of some incorrect detections. To prevent inter-person penetrations in close human reconstruction, we adopt a differentialable 3D distance fields~\cite{Tzionas2016} to reflect the mesh collision: 

\begin{equation}
    \begin{split}
        \mathcal{L}_{\text{pen}} = \sum \limits_{\left(f_a, f_b\right) \in \mathcal{C}} \left\{ \sum_{v_a \in f_a}\left\|-\Psi_{f_b}\left(v_a\right) n_a\right\|_2^2 + \right. \\
        \left. \sum_{v_b \in f_b}\left\|-\Psi_{f_a}\left(v_b\right) n_b\right\|_2^2 \right\}
    \end{split}
    \end{equation}
where $f_a, f_b$ are two colliding triangles in the detected colliding triangles $\mathcal{C}$. $v$ and $n$ are vertex position and normal, respectively, and $\Psi(\cdot)$ is the distance field. We also use a smoothness term to enforce smooth motions:
\begin{equation}
    \mathcal{L}_{\text{smooth}} = \sum_{i=1}^{N-1} \|J^{i+1}_{3D} - J^i_{3D}\|_2^2.
\end{equation}
We further regularize the predicted motion and appearance parameters by:
\begin{equation}
    \mathcal{L}_{\text{reg}} = \|\theta - \theta^{\prime}\|_2^2 + \|\tau - \tau^{\prime}\|_2^2 +  \|\beta - \beta^{\prime}\|_2^2 + \mathcal{L}_{\text{offset}} + \mathcal{L}_{\text{scale}},
\end{equation}
where $\theta^{\prime}$, $\beta^{\prime}$, and $\tau^{\prime}$ are the initial predictions from the diffusion model. $\mathcal{L}_{\text{offset}} = \|\mu\|_2^2$ and $\mathcal{L}_{\text{scale}}=\|s\|_2^2$ calculate the L2-norm of the predicted offsets and scales, respectively.

\vspace{1mm}
\noindent\textbf{Joint optimization for motion and appearance.}
Given an in-the-wild video with two-person close interactions, we first track each human with AutoTrackAnything~\cite{AutoTrackAnything} to produce bounding-boxes and masks. VitPose~\cite{xu2023vitpose} is also used to detect 2D keypoints for each human. 
To be noted that other methods require precise segmentation or keypoints of each individual for reconstruction. In contrast, our framework simultaneously splats two interactive humans and requires only the segmentation of all individuals as a whole to mask the background, which avoids the individual semantic parsing in close interactions. Subsequently, we predict the initial motions with the trained proxemics prior. After the initialization, the overall optimization objective is defined as:
\begin{equation}
    \argmin_{\pi_{m}, \pi_{a}} \mathcal{L}=\mathcal{L}_{\text{app}} + \mathcal{L}_{\text{reproj}} + \mathcal{L}_{\text{pen}} + \mathcal{L}_{\text{smooth}} + \mathcal{L}_{\text{reg}}.
\end{equation}
The optimization variables are $\pi_{m}$ and $\pi_{a}$, where $\pi_{a}$ represents the parameters of the optimizable tensors and U-Net in the appearance branch. We adopt the Adam optimizer with learning rates of 0.00002 and 0.003 for the motion and appearance branches, respectively. The optimization typically takes 
$\sim3-5$ minutes for a video with 128 frames.

%% file: sec/4_dataset.tex
\section{WildCHI Dataset}\label{sec:Dataset}
\begin{table*}
    \begin{center}
        \resizebox{0.9\linewidth}{!}{
            \begin{tabular}{l c c c c c c c c}
            \noalign{\hrule height 1.5pt}
            \begin{tabular}[l]{l}\multirow{1}{*}{Dataset}\end{tabular}
                &Motions &Frames &Scene &3D Pose Format &Scheme &Temporal &RGB image\\
            \noalign{\hrule height 1pt}
            CHI3D~\cite{fieraru2020three}       &373 &63K &indoor     &SMPLX    &MoCap   &\ding{51} &\ding{51} \\
            Hi4D~\cite{yin2023hi4d}        &100 &11K &indoor     &SMPL     &mRGB    &\ding{51} &\ding{51}\\
            ExPI~\cite{guo2022multi}        &115 &30K &indoor     &Skeleton &mRGB    &\ding{51} &\ding{51}\\
            InterHuman~\cite{liang2023intergen}  &6,022 &1.7M &indoor  &SMPL     &mRGB    &\ding{51} &\ding{55} \\
            Inter-X~\cite{xu2023inter}     &11,388 &8.1M &indoor &SMPLX    &MoCap   &\ding{51} &\ding{55} \\
            Flickr Fits~\cite{muller2023generative} &-- &11K &outdoor     &SMPLX    &Pseudo  &\ding{55} &\ding{51}\\
            \hline
            \textbf{WildCHI} &100 &20K &outdoor &SMPL    &Pseudo  &\ding{51} &\ding{51} \\
            \noalign{\hrule height 1.5pt}
            \end{tabular}
        }
        \vspace{-2mm}
        \caption{\textbf{Comparisons of existing human-human interaction datasets.}}
\label{tab:dataset}
\end{center}
\vspace{-9mm}
\end{table*}

Despite the prosperous development of 3D human datasets in recent years, data for close human interaction remains scarce due to complex body contacts, extreme inter-person occlusions, and severe visual ambiguities. Although some recent works have introduced several large-scale human-human interaction datasets~\cite{liang2023intergen,xu2023inter}, they lack RGB images and cannot be used for the human interaction reconstruction task. Furthermore, other datasets are captured in controlled environments, leading to a domain gap from in-the-wild images. The only outdoor dataset available is Flickr Fits~\cite{muller2023generative}, which annotates images with close interactions under a contact constraint. However, no in-the-wild dataset exists to learn temporal dependencies for close human interactions, which are crucial for addressing occlusions and depth ambiguities. To this end, we collect 100 videos with diverse environments and subjects from TikTok~\cite{tiktok}, and build pseudo ground-truth with the proposed method. We also manually filter out incorrect estimations. \cref{tab:dataset} demonstrates the strengths of our in-the-wild close human interaction~(\textbf{WildCHI}) dataset, which has a similar amount of motion as commonly used indoor datasets (Hi4D and ExPI). We also train CloseInt~\cite{huang2024closely}, a regression-based close human interaction method, on the proposed dataset. The experimental results in \cref{tab:comparison} show that WildCHI can improve the performance of CloseInt~\cite{huang2024closely} in both indoor and outdoor scenarios. According to the terms of service of TikTok~\cite{TikTok_term}, we will release our dataset for research purposes. More animatable samples of the proposed dataset can be found in the Supplementary Material.

%% file: sec/5_experiments.tex
\section{Experiments}\label{sec:Experiments}

\subsection{Datasets}\label{sec:datasets}
\noindent\textbf{Inter-X~\cite{xu2023inter}} and \textbf{InterHuman~\cite{liang2023intergen}} are large-scale human-human interaction datasets. Inter-X covers 40 daily interaction categories with 89 distinct subjects having different social relationships. InterHuman also contains diverse two-person interactions. Due to the lack of color images, we use these datasets to train the proxemixs prior only.
\noindent\textbf{Hi4D~\cite{yin2023hi4d}} is an accurate multi-view dataset capturing closely interacting humans. It contains 20 unique pairs of participants with varying body shapes and clothing styles performing diverse interaction motion sequences. We follow \cite{huang2024closely} to use 5 pairs~(23, 27, 28, 32, 37) as testset. The remaining sequences are used for training.
\noindent\textbf{CHI3D~\cite{fieraru2020three}} captures 3 pairs of people in close interaction scenarios with a Vicon MoCap system and 4 additional RGB cameras. We use the standard splits of this dataset.
\noindent\textbf{3DPW~\cite{vonMarcard2018}} also contains several sequences with two-person interactions. We use these sequences for evaluation only.

\subsection{Metrics}\label{sec:Metrics}

We report Mean Per Joint Position Error (MPJPE) and MPJPE after Procrustes Analysis (PA-MPJPE) on close human interaction datasets. The joint PA-MPJPE~\cite{muller2023generative} is also used, which applies Procrustes Analysis on the pair. Additionally, we utilize the Mean Per Vertex Position Error (MPVPE) to measure mesh quality. Following \cite{huang2024closely}, we use an interaction error to assess the quality of reconstructed interactive behaviors. Moreover, we incorporate the average penetration depth (A-PD) \cite{rong2021monocular} to evaluate body contact and penetration depth.

\begin{figure*}
    \begin{center}
    \includegraphics[width=0.9\linewidth]{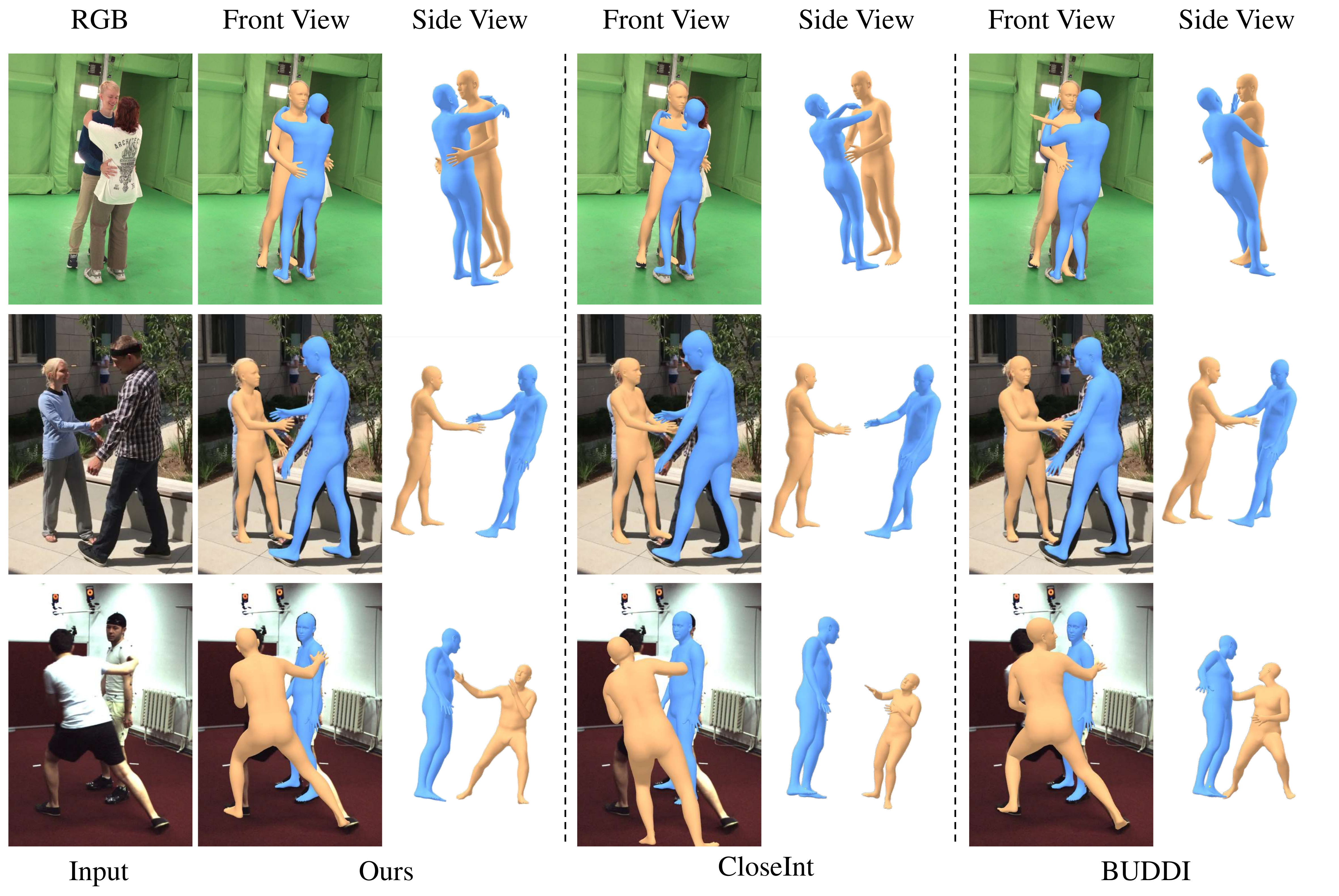}
    \end{center}
    \vspace{-7mm}
    \caption{Qualitative comparison with BUDDI~\cite{muller2023generative} and CloseInt~\cite{huang2024closely}. Our method is more robust to visual ambiguity.}
\label{fig:comparison}
\vspace{-2mm}
\end{figure*}

\begin{table*}
    \begin{center}
        \resizebox{0.9\linewidth}{!}{
            \begin{tabular}{l|c c c c|c c c c}
            \noalign{\hrule height 1.5pt}
            \begin{tabular}[l]{l}\multirow{2}{*}{Method}\end{tabular}
                &\multicolumn{4}{c|}{Hi4D} &\multicolumn{4}{c}{3DPW} \\
                &MPJPE &PA-MPJPE &MPVPE &Inter &MPJPE &PA-MPJPE &MPVPE &Inter\\
            \noalign{\hrule height 1pt}
            Human4D~\cite{goel2023humans}   &72.1        &52.4            &88.6            &--  &72.9        &49.1            &81.8            &--    \\
            CLLIF~\cite{li2022cliff}   &91.3        &53.6            &109.6            &141.5  &--  &--  &--  &--    \\
            BEV~\cite{sun2022putting}   &91.8        &52.2            &101.2            &131.0  &78.3        &48.5            &82.3            &136.4    \\
            GroupRec~\cite{huang2023reconstructing}   &82.4        &51.6       &88.6         &98.8   &73.3        &48.7            &81.2            &110.6    \\
            BUDDI~\cite{muller2023generative}   &96.8        &70.6            &116.0            &102.6    &83.6        &53.6       &93.8            &113.1      \\
            CloseInt~\cite{huang2024closely}  &63.1 &47.5 &76.4 &81.4  &70.6 &51.4 &80.6 &100.3 \\
            \hline
            CloseInt~\cite{huang2024closely} w/ WildCHI  &61.4 &45.1 &75.4 &80.5 &66.4 & 48.3 &77.4 &\textbf{95.9} \\
            \textbf{Ours} &\textbf{59.1}  &\textbf{44.3}   &\textbf{72.0}  &\textbf{80.2}  &\textbf{64.5}  &\textbf{45.6}   &\textbf{75.2}  &96.4  \\
            \noalign{\hrule height 1.5pt}
            \end{tabular}
        }
        \vspace{-2mm}
        \caption{\textbf{Comparisons on Hi4D and 3DPW.} Our method can achieve state-of-the-art performance in both indoor and outdoor scenarios. ``--" means the results are not available.}
\label{tab:comparison}
\end{center}
\vspace{-9mm}
\end{table*}

\subsection{Comparison to State-of-the-Art Methods}\label{sec:Comparison}
 We compare the proposed dual-branch optimization framework with some state-of-the-art baseline methods on different datasets to demonstrate our superiority. We first evaluate Human4D~\cite{goel2023humans} and CLIFF~\cite{li2022cliff} on Hi4D~\cite{yin2023hi4d} and 3DPW~\cite{vonMarcard2018}. These two methods are designed for single-person scenarios, which directly regress SMPL parameters from a single image. As shown in \cref{tab:comparison}, Human4D~\cite{goel2023humans} achieves high joint accuracy with a ViT backbone but struggles to produce accurate spatial distributions due to the using of weak-perspective camera. On the other hand, CLIFF~\cite{li2022cliff} estimates each human in original camera coordinates but still encounters a high interaction loss due to depth ambiguity. 
 We also compare with BEV~\cite{sun2022putting} and GroupRec~\cite{huang2023reconstructing}, which explicitly consider multi-person scenarios by introducing constraints for crowds. While these methods predict humans with more reasonable distributions, they tend to ignore close interactions.
 
 BUDDI~\cite{muller2023generative} is a recent work designed specifically for close interaction reconstruction using an optimization-based framework. It fits two SMPL models to detected 2D keypoints and can handle in-the-wild images. However, the current state-of-the-art pose detectors~\cite{xu2023vitpose} still struggle to produce reliable keypoints for close interactive cases due to visual ambiguity. Although it significantly improves the pose accuracy, the model training relies on a lot of high-quality interaction data. Since these data can only be obtained in a controlled environment, the trained model shows poor generalization ability in outdoor scenarios. Our approach differs from the above works as we simultaneously reconstruct human motions and appearances, directly utilizing the RGB image as a constraint. This strategy alleviates visual ambiguity by comparing rendered and original images. As depicted in \cref{fig:comparison}, our method estimates interactions with more accurate body poses, depth ordinal relationships, and model-image alignment.

We also conduct experiments on outdoor images. On 3DPW dataset, we follow CloseInt~\cite{huang2024closely} to use all interactive sequences as a benchmark for the evaluation. \cref{tab:comparison} shows that our method can achieve state-of-the-art in terms of most metrics, which demonstrates the superiority of our method in in-the-wild scenarios. As shown in \cref{fig:comparison}, our method can work well under diverse environments. In addition, the close interaction data produced by our method can also significantly improve the current regression-based method~(\eg, CloseInt). We also show some qualitative images and videos in Supplementary Material, which also demonstrate the effectiveness of our method.

\begin{figure*}
    \begin{center}
    \includegraphics[width=1.0\linewidth]{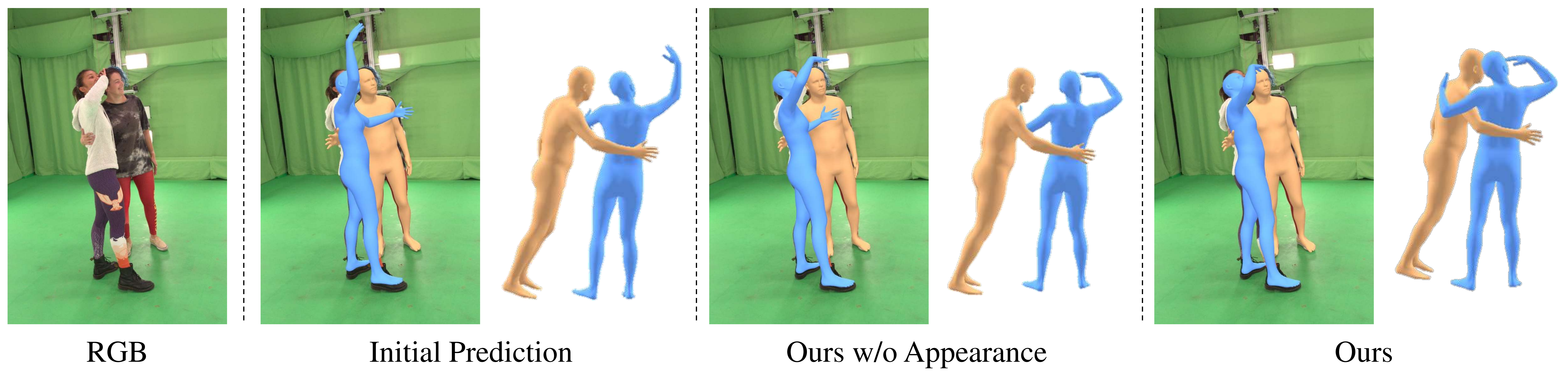}
    \end{center}
    \vspace{-6mm}
    \caption{Ablation study. The initial prediction is severely affected by visual ambiguity and cannot reconstruct accurate interaction. With the proposed optimization, the body pose can be improved with the additional constraints. In addition, we find that appearance constraint is important for the depth ordinal relationships.}
\label{fig:ablation}
\vspace{-5mm}
\end{figure*}

\subsection{Ablation Study}\label{sec:Ablation}
\noindent\textbf{Human appearance.} 
We investigate the importance of the proposed appearance constraint by removing the appearance branch, and supervise the optimization by only motion-level loss functions. As shown in \cref{fig:ablation}, although the framework without the appearance constraint can still produce accurate body poses, the depth ordinal relationship is incorrect. We find that human appearance is effective to prevent this problem, even with coarse texture. 
We simultaneously splat the two-person Gaussians onto the 2D image during human appearance reconstruction to reflect occlusion relationships in the results.
By comparing the rendered image with the original RGB input, we can reason the depth ordinal relationship and enforce better interactions. 
Direct use of RGB images as a constraint also promotes more accurate body poses and model-image alignment since it does not introduce noises compared to 2D keypoints and masks.
Additionally, physical constraints always limit the solution space of the optimization~\cite{shimada2020physcap}. When the physical constraint is not applied, the appearance and proxemics loss promote the optimization to find more accurate 3D joints without considering the mesh penetration.

\noindent\textbf{Proxemics prior.} 
The primary limitations of optimization-based human reconstruction are the sensitivity to depth ambiguity and local minima. To alleviate the impact of these two obstacles, we propose a proxemics prior learned from extensive interaction data to assist the optimization. We formulate this prior as a diffusion model, allowing it to denoise noisy motions and generate clean data. During the optimization, we finetune the pretrained network parameters, and enforce the network to output accurate motions under various supervisions. In \cref{tab:ablation}, we compare our approach with a strategy that directly optimizes SMPL parameters without the proxemics prior. By leveraging learned network parameters containing pose and interaction prior knowledge, optimization with the prior proves to be more accurate and efficient.

\begin{table}
    \begin{center}
        \resizebox{1.0\linewidth}{!}{
            \begin{tabular}{l|c c c c c}
            \noalign{\hrule height 1.5pt}
            \begin{tabular}[l]{l}\multirow{1}{*}{Method}\end{tabular}
                &MPJPE &PA-MPJPE &MPVPE &Interaction &A-PD \\
            \noalign{\hrule height 1pt}
            Initial Prediction        &65.05      &48.54      &78.35      &86.20  &1.16 \\
            Ours w/o Appearance       &60.68      &45.86      &73.52      &81.01  &0.83\\
            Ours w/o Proxemics        &61.52      &47.13      &74.84      &87.13  &0.85\\
            Ours w/o Physics          &57.01      &42.67      &69.57      &78.50  &1.30 \\
            Ours                      &59.06      &44.29      &71.99      &80.18  &0.81 \\
            \noalign{\hrule height 1.5pt}
            \end{tabular}
        }
        \vspace{-2mm}
        \caption{\textbf{Ablations on Hi4D.} "Initial Prediction" denotes the results directly predicted by the pretrained proxemics prior without the optimization. "w/o Appearance", "w/o Proxemics" and "w/o Physics" represent the frameworks without appearance constraint, proxemics prior, and physical constraint, respectively.}
\label{tab:ablation}
\end{center}
\vspace{-8.5mm}
\end{table}

%% file: sec/6_conclusion.tex
\vspace{-1mm}
\section{Limitation and Future Work}\label{sec:Limitation}
\paragraph{Limitation.} Although the proposed framework can produce 3D close human interactions from a monocular in-the-wild video, there are still some limitations. First, our method cannot reconstruct high-quality complete human textures when the light condition is changed or the human is partially observed. Although a coarse texture is sufficient for constraining the underlying body motion, the quality of reconstructed appearances can still be improved by incorporating light embedding~\cite{lin2024vastgaussian} or large vision foundation models~\cite{lee2024guess}. Second, the current design can only capture two-person close interactions. Without sufficient data, we cannot train the proxemics prior for interactive behaviours with more than 2 people. As a result, building an interaction dataset for crowd is also a promising direction for future human behaviour understanding related tasks. In addition, the input video should contain some frames with little or no contact for constraining appearances.

\section{Conclusion}\label{sec:Conclusion}
We propose a novel dual-branch optimization framework to reconstruct two-person close interactions from a monocular in-the-wild video. To alleviate the depth ambiguity and insufficient visual information, we first introduce a proxemics prior based on diffusion model to assist the optimization. We then build a appearance branch with 3D Gaussian splatting to address the notorious visual ambiguity. Compared to previous works that rely on keypoints, masks, and pure RGB information, our method is more robust to diverse environments and can produce more accurate results. Based on the propose framework, we further build an in-the-wild close interaction dataset to promote related research.

\paragraph{Acknowledgement.} 
This research / project is supported by the National Research Foundation (NRF)
Singapore, under its NRF-Investigatorship Programme (Award ID. NRF-NRFI09-0008), and the China Scholarship Council under Grant Number 202306090192.